\documentclass[10pt,twocolumn,letterpaper]{article}

\usepackage{cvpr}              %

\usepackage{capt-of}
\usepackage[dvipsnames]{xcolor}
\usepackage{xspace}
\usepackage{enumitem}
\usepackage{amsmath,amssymb,amsbsy,amsfonts,dsfont,pifont,bm,bbm,mathrsfs,mathtools,nicefrac}
\usepackage{algorithm,algpseudocode,listings}
\usepackage{booktabs,multirow,adjustbox,diagbox,threeparttable,tabularray}
\usepackage[symbol*]{footmisc} %

\newcommand{\methodname}{ScaleLSD}
\newcommand{\method}{\texttt{\methodname}\xspace}

\definecolor{cvprblue}{rgb}{0.21,0.49,0.74}
\usepackage[pagebackref,breaklinks,colorlinks,citecolor=cvprblue]{hyperref}
\usepackage{subcaption}
\usepackage{wrapfig}
\usepackage[capitalize]{cleveref}  %
\crefname{section}{Sec.}{Secs.}
\Crefname{section}{Section}{Sections}
\crefname{table}{Tab.}{Tabs.}
\Crefname{table}{Table}{Tables}
\crefname{figure}{Fig.}{Figs.}
\Crefname{figure}{Figure}{Figures}
\crefname{equation}{Eq.}{Eqs.}
\Crefname{equation}{Equation}{Equations}
\usepackage[markup=underlined]{changes} %

\title{ScaleLSD: Scalable Deep Line Segment Detection Streamlined}

\author{Zeran Ke$^{1,2}$
\qquad  Bin Tan$^2$
\qquad Xianwei Zheng$^1$
\qquad Yujun Shen$^2$
\qquad Tianfu Wu$^4$
\qquad Nan Xue$^{\dag 2}$\\
$^1$Wuhan University \quad 
$^2$Ant Group \quad 
$^3$NC State University
}

\begin{document}
\twocolumn[{%
\renewcommand\twocolumn[1][]{#1}
\maketitle
\begin{center}
    \centering
    \includegraphics[width=.96\linewidth]{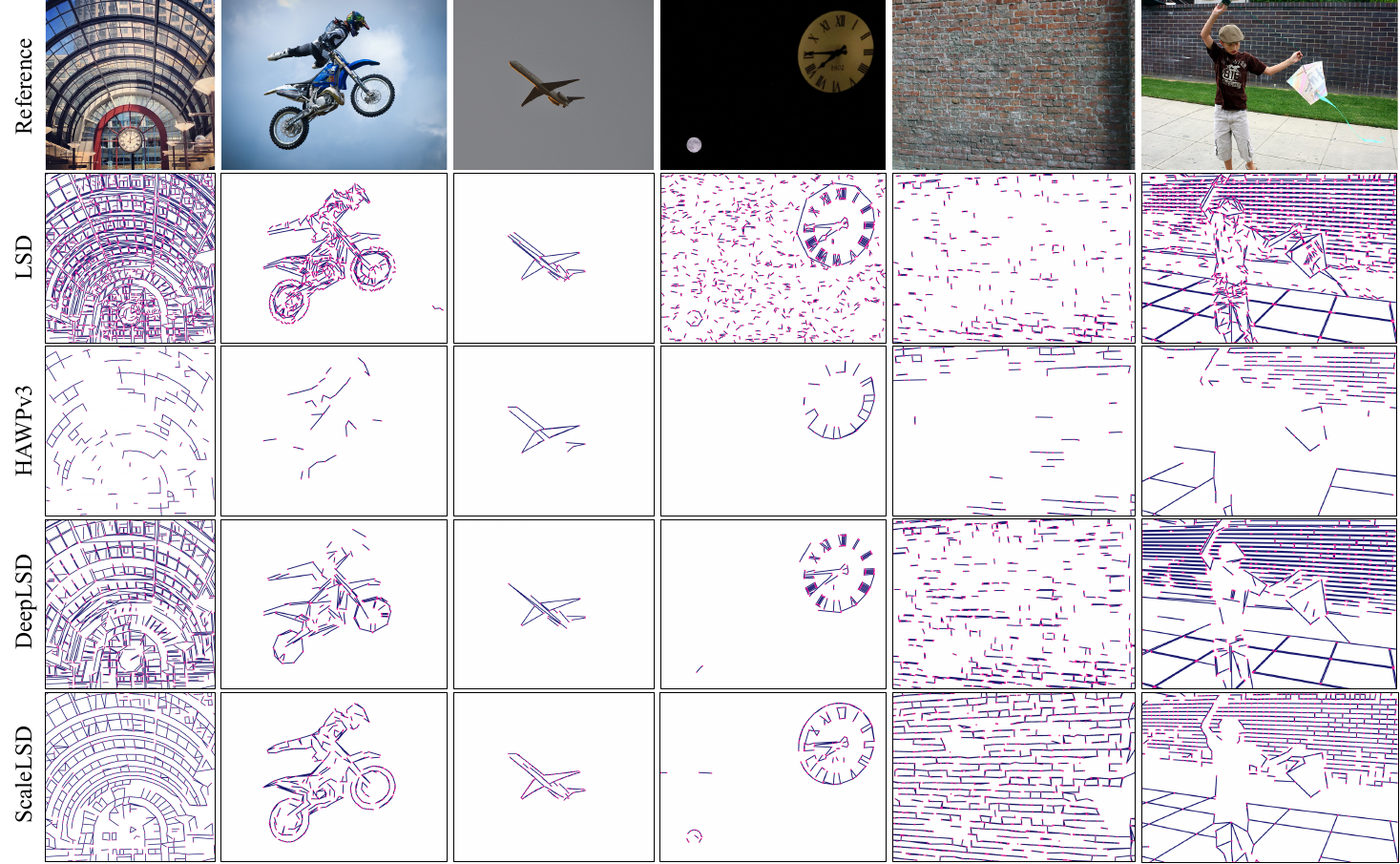}
\end{center}
\vspace{-6mm}
\captionof{figure}{Our ScaleLSD handles a wide range of images, depicting their geometric structures (including the curves, object contours, repeated patterns, and structural regularities) by self-supervised learning of line segment detection from 10M unlabeled images. 
}
\label{fig:teaser}
\vspace{3mm}
}]
\footnotetext[2]{Corresponding author.}

\setlength{\abovecaptionskip}{1pt}
\setlength{\belowcaptionskip}{1pt}
\setlength{\belowdisplayskip}{1pt} \setlength{\belowdisplayshortskip}{1pt}
\setlength{\abovedisplayskip}{1pt} \setlength{\abovedisplayshortskip}{1pt} 

\begin{abstract}
This paper studies the problem of Line Segment Detection (LSD) for the characterization of line geometry in images, with the aim of learning a domain-agnostic robust LSD model that works well for any natural images. 
With the focus of scalable self-supervised learning of LSD, we revisit and streamline the fundamental designs of (deep and non-deep) LSD approaches to have a high-performing and efficient LSD learner, dubbed as \method, for the curation of line geometry at scale from over $10M$ unlabeled real-world images. 
Our \method works very well to detect much more number of line segments from any natural images even than the pioneered non-deep LSD approach, having a more complete and accurate geometric characterization of images using line segments.
Experimentally, our proposed \method is comprehensively testified under zero-shot protocols in detection performance, single-view 3D geometry estimation, two-view line segment matching, and multiview 3D line mapping, all with excellent performance obtained. 
Based on the thorough evaluation, 
our \method is observed to be the first deep approach that outperforms the pioneered non-deep LSD in all aspects we have tested, significantly expanding and reinforcing the versatility of the line geometry of images.
Code and Models are available at \url{https://github.com/ant-research/scalelsd}
\end{abstract}

\section{Introduction}\label{sec:intro}
Boundaries are among the most versatile elements in natural images, as low-complexity composable primitives to depict the complicated geometric shapes~\cite{LiLM20}, their spatial and topological relationships~\cite{PatchLine}, as well as the shape-related high-level semantics~\cite{line3dpp,limap,neat,DuanL15,YueKSE23,LazarowXT22,XiaHXLZ19,abs-2208-00609} and semantic structures in natural scenes~\cite{Sem_Lines_eccv2020,Sem_Lines_TPAMI2021}. 
There has been a vast body of literature~\cite{HarrisS88,MikolajczykS04,XueXBZS18,XiaDG14,bsds,HED-ijcv,CATS,Ballard81,VonGioi2010} on the computational characterization of boundaries in images at different levels of representation (including corner points, edges, line segments, curves, and contours), first via directly modeling image gradients for a long period, and then transitioned into learning paradigms empowered by deep neural networks and (labor-intensive) annotated datasets. In this paper, we are interested in line segment detection for the geometric characterization of images (Fig.~\ref{fig:teaser}), which is useful for many downstream 3D vision tasks due to the parsimoniousness and expressivity of line segments.

Recent studies of deep learning based LSD have been largely driven by meticulously annotated line segments of the Wireframe dataset~\cite{wireframe-dataset}. 
Featured by their non-local and vectorized boundary structures, the 5K training data from the Wireframe dataset have enabled and spurred the development of deep line segment detectors in supervised learning settings~\cite{AFM-CVPR,TP-LSD,LETR,L-CNN,HAWP,HAWPv3}, often with goals to address the locality issue remained in the classical LSD~\cite{VonGioi2010}. 
However, these supervised learning LSD methods struggled with limited generalization to natural images in the wild, which not be easily addressed via scaling up due to the label-intensive and error-prone process of annotating line segments in natural images. 
Self-supervised learning (SSL) approaches for LSD~\cite{SOLD2,DeepLSD,HAWPv3} underscored the limitations of human-annotated labels, and improved the generalizability of LSD over fully supervised counterparts. Nevertheless, the classical LSD~\cite{VonGioi2010} often has a higher recall rate than SSL LSD approaches~\cite{DeepLSD,HAWPv3}.  

\textbf{Our aim} in this paper is to devise a method capable of autonomously ``defining" boundary line geometries by harnessing image data at scale, to tackle the generalization issue in self-supervised learning of LSD.
We hypothesize that existing self-supervised LSD approaches might be limited mainly by their training scales using only thousands of images, but realize that the automatic labeling pipelines in state-of-the-art SSL approaches for LSD, HAWPv3~\cite{HAWPv3}, SOLD$^2$~\cite{SOLD2} and DeepLSD~\cite{DeepLSD}, \textit{have scalability issues}. 

In both HAWPv3~\cite{HAWPv3} and SOLD$^2$~\cite{SOLD2}, the homographic adaptation schema in the labeling pipeline often leads to low recall rates of line segments in unlabeled images, prohibiting effective large-scale SSL that entails sufficient high-quality pseudo labels. 
To address the low recall rate issue, DeepLSD~\cite{DeepLSD} exploits the local meaningful alignment schema proposed in the classical  LSD~\cite{VonGioi2010} in the pseudo-label generation, but
unavoidably inherits its locality issue, and experimentally converges to the performance of the classical LSD~\cite{VonGioi2010} for downstream tasks as we shall show in experiments (see Fig.~\ref{fig:teaser} for qualitative comparisons). 

Scalability entails simplicity in SSL with large-scale unlabeled data, as witnessed by the recent unprecedented progress made in natural language understanding and mid-to-high-level computer vision tasks in the literature. After carefully revisiting state-of-the-art SSL based LSD approaches~\cite{SOLD2,HAWPv3,DeepLSD} with the simplicity principle in mind, \textbf{we streamline those approaches with three key observations and design choices highlighted}: 
\begin{itemize}
    \item The holistic attraction (HAT) field representations~\cite{HAWP,AFM-CVPR,HAWPv3} have great potential in SSL of LSD, and predicting the HAT field precisely from images can simplify the LSD modeling. 
    \item Image attributes, inductive biases, and meticulous designs of the classical LSD~\cite{VonGioi2010} facilitate the self-supervised learning by inducing a super-efficient and high-recall pseudo labeling pipeline, working well in the integration with the HAT field learning at scale.
    \item Expressive Transformer~\cite{vaswani2017attention} based backbones are critical for ``ingesting" large-scale data. 
    \vspace{2pt}
\end{itemize}

We present the \method method, which works well for any natural images after training with 10M unlabeled images sourced from the SAM-1B dataset~\cite{SAM} (Fig.~\ref{fig:teaser}). 
 In our experiments,  we showcase the final model of our \method significantly advanced detection performance measured by the repeatability rate on several data collections that are all different from the training distribution. 
We further demonstrate that a comprehensive and accurate characterization of line geometry facilitates all the downstream 3D vision tasks, of single-view vanishing point estimation, two-view line segment matching, and multiview 3D line reconstruction, all obtaining state-of-the-art performance.

\section{Related Work}\label{sec:related}
Traditional handcraft line segment detection methods~\cite{LSD,EDLines,ELSED} primarily rely on low-level image feature processing.
Transformer-based fully supervised methods~\cite{LETR,HEAT} eliminate traditional edge detection steps and directly regress line endpoints using a Transformer decoder.
For the deep learning based approaches, the focus of LSD has been shifted from fully supervised learning~\cite{AFM-CVPR,RegionalAttraction,PPGNet,L-CNN,HAWP,HAWPv3,LGNN,TP-LSD,FClip,gu2021realtime,ELSD,LETR,DeepHoughPrior}  on the Wireframe dataset~\cite{wireframe-dataset} to self-supervised approaches to address generalization issues of deep LSD models. 

\noindent\textbf{Self-supervised LSD Learning.}
The development of self-supervised LSD learning revealed that the human-annotated line geometry in real-world images contains biases, often leading to suboptimal performance in downstream 3D vision tasks such as vanishing point estimation~\cite{Denis2008} and multi-view 3D line reconstruction~\cite{limap, neat, line3dpp}. 
SOLD$^2$~\cite{SOLD2} presented the first automatic line geometry labeling process, which took advantage of the inherent generalization ability of boundaries to annotate line segments in a sim-to-real pipeline, in which the homographic adaptation scheme was shown to be useful to eliminate erroneous detection results by averaging multiple inference results up to random homographic warping of unlabeled images. Follow-up studies improved the efficiency and effectiveness of homographic adaptation for better self-supervised learning models~\cite{HAWPv3,DeepLSD,L2D2}.  In our study, we found, the cost of homographic adaptation schema for erroneous detection filtering is the completeness during the pseudo label generation for large-scale data, which limits the self-supervised learning of LSD at a small-scale scenario. Our presented method further demonstrate that the homographic adaptation scheme is not necessary for better self-supervised learning of LSD.

\noindent\textbf{Attraction Field Representations.}
The recent self-supervised LSD methods~\cite{DeepLSD,HAWPv3} were benefited from attraction field representations~\cite{AFM-CVPR,HAWP} that parameterize sparse line segments using dense fields. DeepLSD~\cite{DeepLSD} further demonstrated that the classic LSD approach~\cite{VonGioi2010} facilitates self-supervised LSD learning, but it extensively relies on the local alignment scheme proposed in the LSD~\cite{VonGioi2010}. 
Our proposed work is inspired by DeepLSD~\cite{DeepLSD}, but finds a different role of the classic LSD in self-supervised learning, in which LSD~\cite{VonGioi2010} is leveraged for rectifying prediction errors during the learning of holistic attraction fields, allowing large-scale self-supervised learning of LSD.

\section{Approach}\label{sec:method}
In this section, we first present background on the HAT field representation~\cite{HAWP,HAWPv3} and the direction / level-line field in the classica LSD~\cite{VonGioi2010} to be self-contained. We then present details of our streamlined formulation of \method (~\cref{fig:network}) on top of HAWPv3~\cite{HAWPv3}, followed by details of pseudo line segment label generation (~\cref{fig:data-pipeline}). 

\subsection{Background on Line Segment Representation}

\begin{wrapfigure}{r}{0.45\linewidth}
    \centering
    \includegraphics[width=\linewidth]{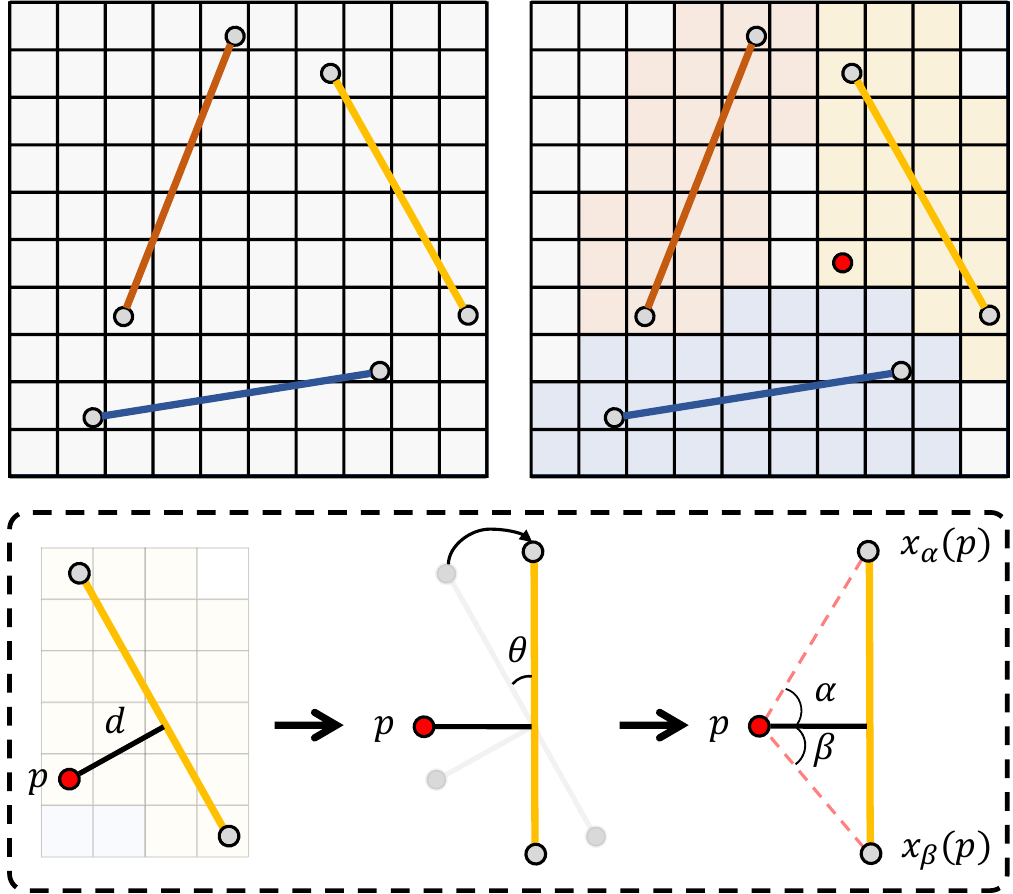}
    \captionof{figure}{Illustration of the HAT field proposed in~\cite{HAWPv3}.}
    \label{fig:hat-field}
\end{wrapfigure}

The HAT field representation~\cite{HAWP,HAWPv3} lifts line segments to attraction regions (~\cref{fig:hat-field}), which depicts the full geometry of the line segment set defined on the discrete image grid using a rather dense number of pixels in a dense representation. Formally, For a set of line segments $\mathcal{L} = \{\ddot{l}_i = (\mathbf{x}_i^{0},\mathbf{x}_i^{1})\}_{i=1}^N$ defined on an $H\times W$ image grid, the HAT field maps the set $\mathcal{L}$ in a 4-component field $\mathcal{H}(\mathbf{p}) = \left(d(\mathbf{p}), \theta(\mathbf{p}), \alpha(\mathbf{p}), \beta(\mathbf{p})\right)$ in which each (foreground) pixel $\mathbf{p}$ is assigned to its perpendicularly closest line segment, where $d(\mathbf{p})\in (0,+\infty)$ and $\theta(\mathbf{p}) \in (-\pi,\pi]$ measure the perpendicular distance and direction of the line segment respectively, and $\alpha(\mathbf{p})\in (-\pi/2,0),\beta(\mathbf{p})\in (0, \pi/2)$ characterize the two vectors pointing from $\mathbf{p}$ to the two endpoints in the local coordinate frame origin at $\mathbf{p}$ with the direction $\theta$ as the $x$-axis. The two endpoints of a line segment $\ddot{l}(\mathbf{p}) = (\mathbf{x}_{\alpha}(\mathbf{p}),\mathbf{x}_{\beta}(\mathbf{p})) \in \mathbb{R}^2\times \mathbb{R}^2$ defined by the pixel $\mathbf{p}$ is computed from the 4D distance-angle parametrization by,

\begin{equation}\label{eq:endpoint-field}
    \ddot{l}(\mathbf{p}) = d \cdot
    \begin{bmatrix}
        \cos \theta & -\sin \theta \\
        \sin \theta & \cos \theta 
    \end{bmatrix}
    \begin{bmatrix}
        1 & 1 \\
        \tan \alpha & \tan \beta
    \end{bmatrix} + 
    \begin{bmatrix}
        \mathbf{p},
        \mathbf{p}
    \end{bmatrix}.
\end{equation}

It is thus straightforward to learn HAT field representation for line segment detection when the ground-truth (GT) label of line segments are available, thus inducing the self-supervised learning of LSD in a pseudo labeling pipeline starting from {a bootstrap training using} synthetic data with clearly defined GT labels. 

\textit{The level-line field in the LSD}~\cite{VonGioi2010} is consistent with the $\theta$ field in the HAT field, which is computed by a well-tailored algorithm, and leveraged in our proposed pseudo labeling pipeline to counter the gap between synthetic images and real images.  

\subsection{The Meta Architecture}
\begin{figure*}[t]
    \centering
    \begin{subfigure}{\linewidth}
    \includegraphics[width=\linewidth]{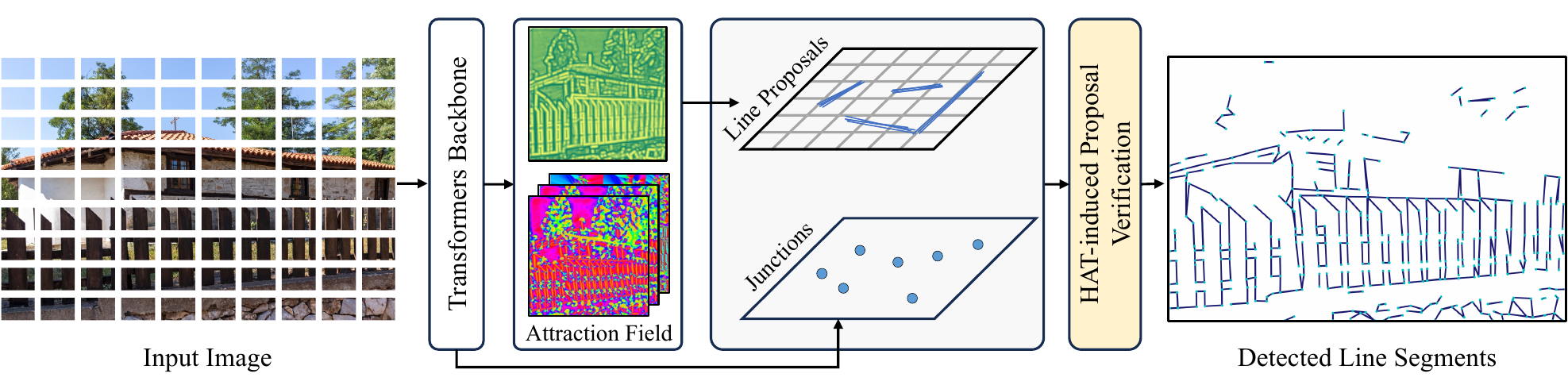} 
    \caption{
    The architecture overview of the proposed \method for line segmentation detection.}
    \vspace{2mm}
    \label{fig:network}
    \end{subfigure}
    \begin{subfigure}{\linewidth}
    \includegraphics[width=1.\linewidth]{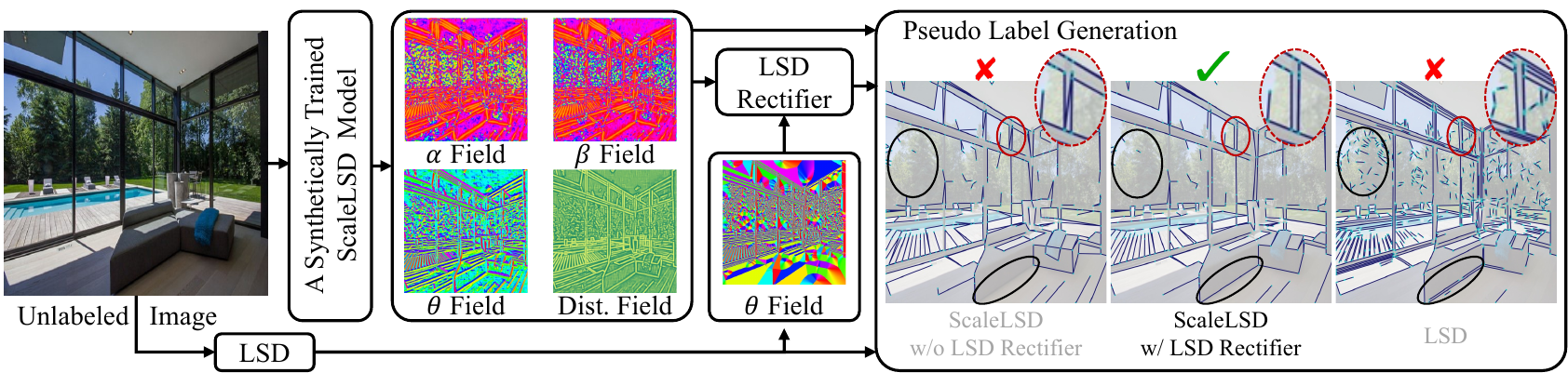} 
    \caption{Illustration of the pseudo label generation pipeline on the real image.}
    \label{fig:data-pipeline}\vspace{3mm}
    \end{subfigure}

    \caption{The network architecture and the pseudo label generation in the proposed \method. Vision Transformer backbones ensure the effectiveness of HAT field learning, thus allowing us to use a HAT-induced verification scheme to decode line segments. In the pseudo label generation, we present an efficient pipeline that use the local line segments by the classical LSD~\cite{VonGioi2010} to rectify the network outputs, enabling the large-scale training of LSD with high-quality pseudo labels.
    }
    \label{fig:method-and-data}
\end{figure*}

\cref{fig:network} illustrates the meta architecture. Compared to HAWPv3~\cite{HAWPv3}, the proposed architecture significantly streamlines designs with a novel method for HAT-induced proposal verification.

\vspace{2mm}
\noindent\textbf{Line Segment Proposal.}
While the HAT field representation has a clear form to represent each line segment in the set $\mathcal{L}$, the dense field representation, together with the uncertainty in the learning process,  often lead to a large number of duplicated proposals for each line segment.
As the goal of LSD is to compute a parsimonious and sparse characterization of input images, it is required to sparsely decode the HAT field. To that end, the junctions/endpoints are learned together with the HAT field, leading to the desired sparse decoding scheme.
Denoted by  $\mathcal{J} = \{\jmath_1,\ldots,\jmath_M\}\subset \mathbb{R}^2$ the set of learned junctions, the sparse decoding scheme first binds the endpoint fields $\mathbf{x}_{\alpha}(\mathbf{p})$ and $\mathbf{x}_{\beta}(\mathbf{p})$ by finding their closest junction, indexing each line segment $(\mathbf{x}_{\alpha}(\mathbf{p}),\mathbf{x}_{\beta}(\mathbf{p}))$ into $(\imath_\alpha(\mathbf{p}),\imath_\beta(\mathbf{p}))$, where the index mapping $\imath_\alpha(\mathbf{p})$ (or $\imath_\beta(\mathbf{p})$) is defined by 

\begin{equation}\label{eq:sparse-decoding}
\begin{split}
    \imath_\alpha(\mathbf{p}) = \arg\min_{j} \|\mathbf{x}_\alpha-\jmath_j\|, %
    \\
    \imath_\beta(\mathbf{p}) = \arg\min_{j} \|\mathbf{x}_\beta-\jmath_j\|,
\end{split}
\end{equation}
$\imath_\alpha,\imath_\beta \in \{0,\ldots, M\}$. 
Note, when $\imath_\alpha$ (or $\imath_\beta$) becomes $0$, it means the minimal distance defined in \cref{eq:sparse-decoding} is larger than a threshold $\tau_{\rm dist}$, which is set to $10$ pixels in our experiments to prune out the outliers in the field prediction. 
 With the index mapping, the line segments in the field with the same index pair (up to the order swapping) are regarded as the same line segment, finally obtaining a sparse set of line segments, each endpoint of which belongs to the set $\mathcal{J}$.  Because the endpoint indices are represented in integers, a GPU-builtin implementation yields the unique line segments (and unique index pairs) with little latency.

\vspace{2mm}
\noindent\textbf{The HAT-Induced Proposal Verification.} \label{subsec:proposal-verification}
 The proposal verification were extensively studied to prune out the false detections from the generated proposals for both the classical LSD approach~\cite{VonGioi2010} in an a-contrario line verification scheme and the learning-based approaches in the LOI (Line-of-Interest) designs~\cite{L-CNN} that learns the confidence score of each line proposal according to the ground-truth labels. While LOI-based verification scheme was prevailing in learning-based approaches for end-to-end learning, it poses an issue of label reliability in self-supervised learning, leading to additional designs such as edge map learning and edge-guided verification, as well as the more costly pseudo-label generation schema used in SOLD$^2$ and HAWPv3.

 \begin{figure}[!h]
    \centering
    \includegraphics[width=.95\linewidth]{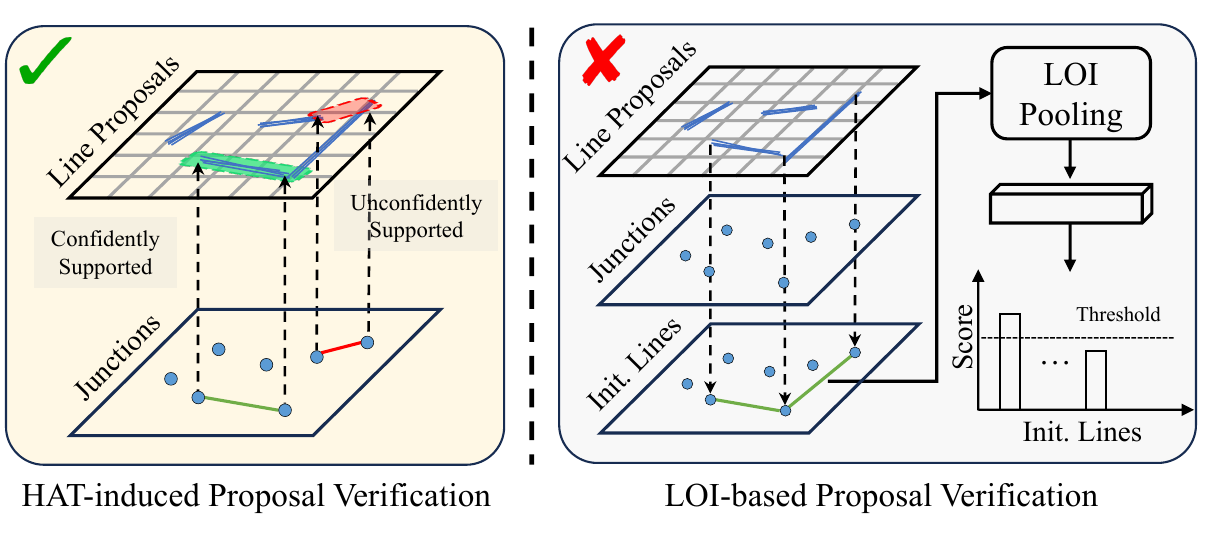}
    \caption{The summary comparison of our proposed HAT-induced proposal verification and the previously used LOI-based proposal verification.}
    \label{fig:verification}
\end{figure}

In our \method, a novel HAT-induced proposal verification is presented, based on the sparse decoding scheme of HAT field in \cref{eq:sparse-decoding}. Denoted by the sparse set of the junction pairs $\mathcal{I} = \{(\imath_{\alpha}^1,\imath_{\beta}^1),\ldots,(\imath_{\alpha}^K,\imath_{\beta}^K)\}$, we check the support degree over the field prediction $(\mathbf{x}_{\alpha}(\mathbf{p}),\mathbf{x}_{\beta}(\mathbf{p}))$ in \cref{eq:endpoint-field} for each index pair $(\imath_{\alpha}^k,\imath_{\beta}^k)$ by 
\begin{equation}
    {\rm Deg}(\imath_{\alpha}^k,\imath_{\beta}^k) = \sum \left((\imath_{\alpha}(\mathbf{p}),\imath_{\beta}(\mathbf{p}))\sim (\imath_{\alpha}^k,\imath_{\beta}^k)\right),
\end{equation}
where $\sim$ operator returns 1 if the two side of inputs are equal up to permutation otherwise 0. By measuring the proposals using the support degree, the larger number of support pixels in the field prediction, the higher the confidence of learned line segments. \cref{fig:verification} made an illustrative comparison between the proposed HAT-induced and the previously-used LOI verification schema. Because the support degree is measured in the number of pixels, it has better explanation than the classification scores by the feature learning, especially in the outlier-contained self-supervised learning with pseudo labels. In our implementations, we default use the threshold of $10$ pixels to filter out the unreliable predictions over the full pipeline of learning.

\subsection{The Pseudo Label Generation in ScaleLSD}
\label{subsec:Pseudo-Label-Generation}
While HAT-induced proposal verification simplified the learning with white-box and geometrically-meaningful designs, we found that the learning of HAT field itself remains problematic, especially in the bootstrapping phase that was trained on the small-scale synthetic data. 
We cope with this issue by delving into the classical design of LSD approach~\cite{VonGioi2010}. We found, although the LSD approach~\cite{VonGioi2010} often produces spurious results in short line segments, its main information source of image gradients are robust and generic to produce reliably line segments when focusing on the orientations, thus bridging the classical design in the learning-based approaches at an appropriate intersection point, leading to an effective design, the LSD-Rectifier for HAT-based self-supervised learning of line segments. 

As shown in \cref{fig:data-pipeline}, given a seed model trained on the synthetic data, we generate pseudo labels on the real-world images by predicting the HAT fields from the seed model as the main source and the LSD approach~\cite{VonGioi2010} as the auxiliary source, then use LSD-Rectifier to replace the $\theta$ component from the main source to the  LSD-sourced one as a rectified HAT field to predict line segments as the pseudo labels. Because the results by the LSD approach~\cite{VonGioi2010} is locally accurate in terms of the line direction, with the proposed LSD-rectifier for pseudo label generation, there is no need to use the computational expensive homographic adaptation schema~\cite{SOLD2,HAWP} to filter out the false detection results. The right of \cref{fig:data-pipeline} qualitatively compared the pseudo labels generated by different schema, showcasing the effectiveness of the LSD Rectifier.

\subsection{Implementation Details}
We adopt the transformer-based architecture (ViT-Base) for feature extraction, and employ the DPT~\cite{DPT} head for HAT field of line segments and 2D heatmap of junctions. 
We maintain the routine of self-supervised learning by the ``synthetic-to-real'' process of training~\cite{SuperPoint,SOLD2,HAWPv3}. The loss functions and training details are provided in the supplementary material.

\vspace{2pt}
\noindent\textbf{Training Datasets.}
Three different datasets are used for training our models.
The synthetic dataset consists of 8 simple primitives and 2k images for each primitive, yields 16k samples for training. 
The Wireframe dataset is augmented by flipping and rotation operations to yield 20k samples for training.
The extensive SA1B dataset contains over 10M images obtained around the world and finally yields over 10M samples for training. See more details in the supplementary material.

\vspace{2pt}
\noindent\textbf{Training Recipes.}
We use the ADAM optimizer \cite{kingma2014adam} for training all models. In the synthetic training stage, we train a preheating model on the synthetic dataset for 10 epochs, and we set the learning rate is initialized as 4e-4 and is divided by 10 at the 7th epoch. 
Then this synthetic model is used to annotate pseudo labels for unlabeled images of realistic dataset. 
In the real training stage, we separately train our model from scratch on the Wireframe dataset for 30 epochs and on the SA1B dataset for 6 epochs. For training a base model on the Wireframe dataset, we set the learning rate is initialized as 4e-4 and is divided by 10 at the 25th epoch. For scaling up on the SA1B dataset, we set the learning rate increases linearly from a base value 2e-4 to a max value 1e-3 in the first 2000 training iterations and then decreases from the max value to the base value in the manner of cosine annealing \cite{SGDR}.

\section{Experiments}\label{sec:exp}
In this section, we evaluate our \method models on four tasks, including detection repeatability, estimation of vanishing points, line segment matching, and 3D line reconstruction. Because our method benefits from large-scale training, the main evaluations are zero-shot. In the final, further analyzes on the HAT-induced proposal verification and pseudo-label generation are reported. For more experimental results, please refer to our supplementary materials.

\begin{table*}[!t]
    \centering
    \resizebox{\linewidth}{!}{
    \begin{tabular}{r|ccccccc|ccccccc}
        \toprule
    \multirow{2}{*}{Method} & \multicolumn{7}{c|}{YorkUrban} & \multicolumn{7}{c}{HPatches} \\
    & Rep-5 (S) $\uparrow$ & Loc-5 (S) $\downarrow$ & Len-5 (S) $\uparrow$ & Rep-5 (O) $\uparrow$ & Loc-5 (O) $\downarrow$ & Len-5 (O) $\uparrow$&	\#Lines/Image& Rep-5 (S) $\uparrow$ & Loc-5 (S) $\downarrow$ & Len-5 (S) $\uparrow$ & Rep-5 (O) $\uparrow$ & Loc-5 (O) $\downarrow$& Len-5 (O) $\uparrow$&	\#Lines/Image\\
    \midrule
    LSD & 0.419 & 2.123 & 0.559 & 0.723 & 0.959 & 0.844 & 591 & 0.275 & 2.673 & 0.264 & 0.424 & 1.779 & \textbf{0.594} & 493 \\
    SOLD$^2$ & 0.585 & 1.918  & 0.548 & \underline{0.824} & 1.097 & 0.803 & 196 & 0.278 & 2.264 & 0.251 & 0.467 & \underline{1.411} & 0.460 & 151 \\
    HAWPv3 & \underline{0.711} & \underline{1.454} & 0.687 & \textbf{0.829} & \underline{0.839} & 0.841 & 225 & 0.322 & \underline{2.314} & 0.317 & 0.509 & 1.572 & 0.528 & 149 \\
    DeepLSD & 0.514 & 2.199 & 0.515 & 0.701 & 1.054 & 0.763 & 310 & 0.241 & 2.548 & 0.228 & 0.457 & 1.894 & 0.493 & 277 \\
    ScaleLSD(Ours)@Wireframe & 0.697 & 1.683 & \underline{0.714} & 0.812 & 0.877 & \underline{0.847} & \underline{598} & \underline{0.337} & 2.318 & \underline{0.348} & \textbf{0.524} & 1.624 & \underline{ 0.567} & \underline{499} \\
    ScaleLSD(Ours)@SA1B & \textbf{0.725} & \textbf{1.265} & \textbf{0.763} & 0.806 & \textbf{0.768} & \textbf{0.849} & \textbf{708} & \textbf{0.367} & \textbf{1.535} & \textbf{0.377} & \underline{0.515} & \textbf{1.187} & 0.549 & \textbf{664} 
    \\\bottomrule
    \multirow{2}{*}{Method} & \multicolumn{7}{c|}{RDNIM} & \multicolumn{7}{c}{COCO Val2017} \\
    & Rep-5 (S) $\uparrow$ & Loc-5 (S) $\downarrow$ & Len-5 (S) $\uparrow$ & Rep-5 (O) $\uparrow$ & Loc-5 (O) $\downarrow$ & Len-5 (O) $\uparrow$ &	\#Lines/Image& Rep-5 (S) $\uparrow$ & Loc-5 (S) $\downarrow$ & Len-5 (S) $\uparrow$ & Rep-5 (O) $\uparrow$ & Loc-5 (O) $\downarrow$& Len-5 (O) $\uparrow$&	\#Lines/Image\\
    \midrule
    LSD & 0.221 & 2.766 & 0.224 & 0.425 & 1.733 & \underline{0.500} & \underline{433} & 0.456 & 2.192 & 0.386 & 0.683 & 1.164 & 0.637 & \underline{561} \\
    SOLD$^2$ & 0.241 & 2.530 & 0.224 & 0.421 & 1.588 & 0.419 & 94 & 0.481 & 2.233 & 0.465 & 0.682 & 0.956 & 0.688 & 83 \\
    HAWPv3 & 0.278 & \textbf{2.200} & 0.268 & 0.420 & \textbf{1.496} & 0.420 & 50 & \underline{0.644} & \underline{1.614} & 0.646 & 0.730 & 1.107 & 0.783 & 99 \\
    DeepLSD & 0.251 & 2.661 & 0.250 & \underline{0.439} & 1.639 & 0.492 & 152 & 0.423 & 2.393 & 0.423 & 0.624 & 1.225 & 0.678 & 207 \\
    ScaleLSD@Wireframe(Ours) & \underline{0.295} & 2.410 & \underline{0.299} & 0.435 & 1.531 & 0.465 & 209 & 0.636 &  1.829 & \underline{0.661} & \underline{0.749} & \underline{0.939} & \underline{0.796} & 346 \\
    ScaleLSD@SA1B(Ours) & \textbf{0.337} & \underline{2.407} & \textbf{0.347} & \textbf{0.491} & \underline{1.510} & \textbf{0.527} & \textbf{540} & \textbf{0.666} & \textbf{1.540} & \textbf{0.699} & \textbf{0.764} & \textbf{0.909} & \textbf{0.809} & \textbf{583}
    \\\bottomrule
    \end{tabular}
    }
    \caption{ The repeatability evaluation results of zero-shot performance on out-of-domain datasets. Numbers with \textbf{bold-font} and \underline{underline} indicate the best and the second best performance on specific metrics. We get the best performance across all datasets and almost all metrics.}
    \vspace{-10pt}
    \label{tab:ood}
\end{table*}

\subsection{Repeatability Scores and Localization Errors}
\paragraph{Datasets and metrics.}
The repeatability scores and localization errors measure the performance of feature detectors for given pairs of images. That is to say, given a pair of images captured for the same thing, we expect a line segment detector to repeatably detect line segments up to the viewpoint or photometric changes. 
We also include the length repeatability evaluation following ELSED~\cite{ELSED}.
Here, 4 datasets, HPatches~\cite{HPatches}, RDNIM~\cite{RDNIM}, YorkUrban~\cite{Denis2008} and COCO (val-2017)~\cite{COCO} are used for the evaluation. For the HPatches~\cite{HPatches} and RDNIM~\cite{RDNIM} datasets, we use the dataset-provided homographies between the image pairs to compute the repeatability scores and localization errors. Because the YorkUrban~\cite{Denis2008} and COCO~\cite{COCO} do not have paired images, we follow the protocol used by previous studies~\cite{DeepLSD,HAWPv3} to warp images by the random homography warping. 
The detection results that are within 5 pixels (in terms of structural distance and orthogonal distance) are regarded as the repeatedly detected line segments for the evaluation. 

\vspace{-10pt}
\paragraph{Baselines.} 
Due to the poor generalization on zero-shot evaluation of supervised methods (e.g., HAWPv1/v2~\cite{HAWP,HAWPv3}, L-CNN~\cite{L-CNN}, etc.), we mainly compare our method with classical LSD~\cite{VonGioi2010} and the leading self-supervised learning approaches SOLD$^2$~\cite{SOLD2}, HAWPv3~\cite{HAWPv3} and DeepLSD~\cite{DeepLSD}. The official implementation and model weights of those approaches are used.
For our \method, two models trained on the Wireframe and SA1B are evaluated. 

\vspace{-12pt}
\paragraph{Results.} 
As reported in ~\cref{tab:ood}, our \method trained on the SA1B data gets the best performance across all out-of-domain datasets and almost all metrics and our base Wireframe model also achieves good results. The classical method LSD and the learning-based combined with LSD method DeepLSD can get comparable performance on the first two datasets, only except that DeepLSD apparently outperforms LSD on the challenging RDNIM dataset but LSD is better than DeepLSD on the COCO Val2017 dataset. HAWPv3 gets lower localization errors than ours on the RDNIM dataset but can only detects a few line segments on all datasets. On the whole, our model has the best and most stable detection capability which is in favor for some downstream tasks of image matching and 3D reconstruction. 

\subsection{Vanishing Points Estimation}
\begin{table}
    \centering
    \resizebox{\linewidth}{!}{
    \begin{tabular}{r|cc|cc}
    \toprule
\multirow{2}{*}{Method} & \multicolumn{2}{c|}{YUD+} & \multicolumn{2}{c}{NYU-VP} \\
& VP Error~$\downarrow$ & AUC~$\uparrow$ & VP Error~$\downarrow$ & AUC~$\uparrow$ \\
\midrule
LSD~\cite{VonGioi2010} & 2.05 & 82.9 (5.3) & 3.29 & 68.6 (6.3) \\
TP-LSD~\cite{TP-LSD} & 1.73 & 85.1 (5.0) & 3.35 & 68.0 (4.5) \\
SOLD$^2$~\cite{SOLD2} & 2.59 & 75.4 (6.4) & 4.46 & 56.9 (7.6) \\
HAWPv3~\cite{HAWPv3} & 1.76 & 84.2 (4.2) & 3.35 & 68.0 (5.7) \\
DeepLSD~\cite{DeepLSD} & 1.63 & 85.6 (3.6) & \underline{3.24} & \underline{69.1} (6.2) \\
ScaleLSD@Wireframe(Ours) & \underline{1.58} & \underline{86.6} (1.9) & 3.81 & 63.9 (3.2) \\
ScaleLSD@SA1B(Ours) & \textbf{1.55} & \textbf{87.1} (1.1) & \textbf{3.18} & \textbf{70.4} (1.4)
\\\bottomrule
    \end{tabular}
    }
    \caption{ Vanishing points estimation on the YUD+~\cite{Denis2008} dataset and the NYU-VP~\cite{NYU-VP} dataset. We make comparisons of all models in term of median VP Error and average AUC (and its standard deviation). }
    \label{tab:vps}
\end{table}
Vanishing points (VP) depict infinity under the projective transformations, and play an important role in single-view 3D geometry. 

\vspace{-10pt}
\paragraph{Baselines.}
We evaluate different line segment detectors (\ie, LSD~\cite{VonGioi2010}, TP-LSD~\cite{TP-LSD}, SOLD$^2$~\cite{SOLD2}, HAWPv3~\cite{HAWPv3}, DeepLSD~\cite{DeepLSD} and our \method) on the VP estimation. We follow DeepLSD~\cite{DeepLSD} to estimate VPs, in which the Progressive-X~\cite{Progressive-X} algorithm is applied to yield vanishing points.

\vspace{-10pt}
\paragraph{Datasets and Evaluation Metrics.}
We use YUD+ and NYU-VP datasets for experiments. YUD+
is extended from the YorkUrban~\cite{Denis2008} dataset and labels up to 8 VPs per image. NYU-VP is adapted from the NYU Depth Dataset V2~\cite{NYU-VP} and labels 1 to 8 VPS per image. 
Two metrics are considered, VP Error measures the precision of the estimated VPs in 3D world by the angular error between the directions of the ground-truth VPs and the predicted VPs. AUC means Area Under the Curve (AUC) of the recall curve of the VPs. 

\vspace{-10pt}
\paragraph{Results.}
As reported in ~\cref{tab:vps}, our base model trained on the structured Wireframe dataset achieves good performance on the YUD+ but drops extremely on the non-Manhattan scenes of the NYU-VP. The scale-up model of our  \method outperforms all baselines in term of VP Error and average AUC (and its standard deviation). 

\begin{figure}
    \centering
    \begin{subfigure}[b]{0.47\linewidth}
        \centering
        \includegraphics[width=\linewidth]{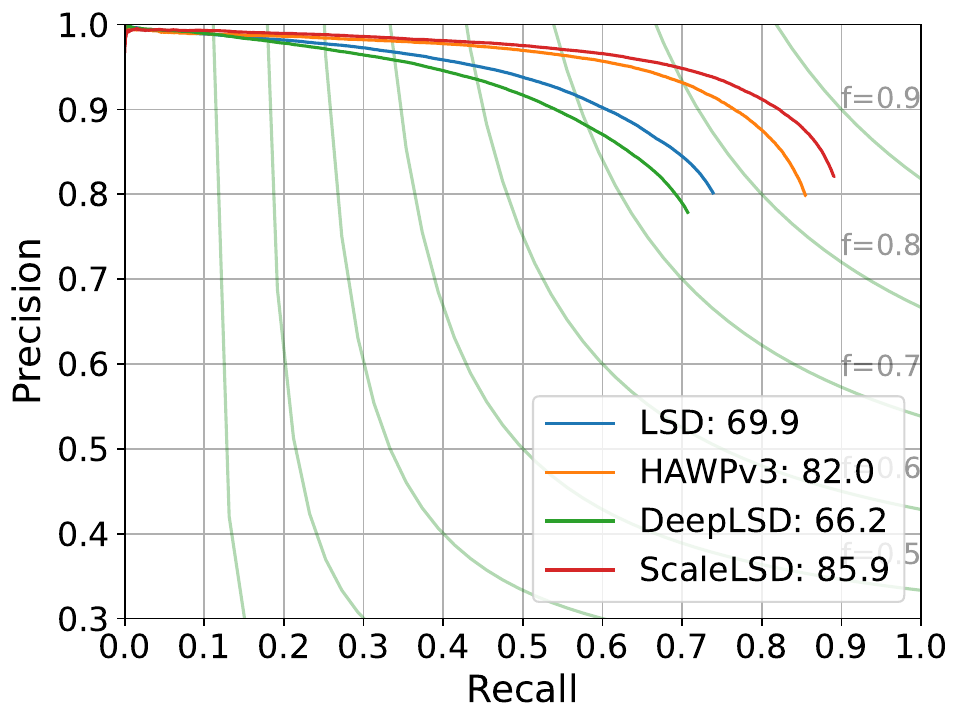}
        \caption{Lines matching}
        \label{fig:line_ap}
    \end{subfigure}
    \begin{subfigure}[b]{0.47\linewidth}
        \centering
        \includegraphics[width=\linewidth]{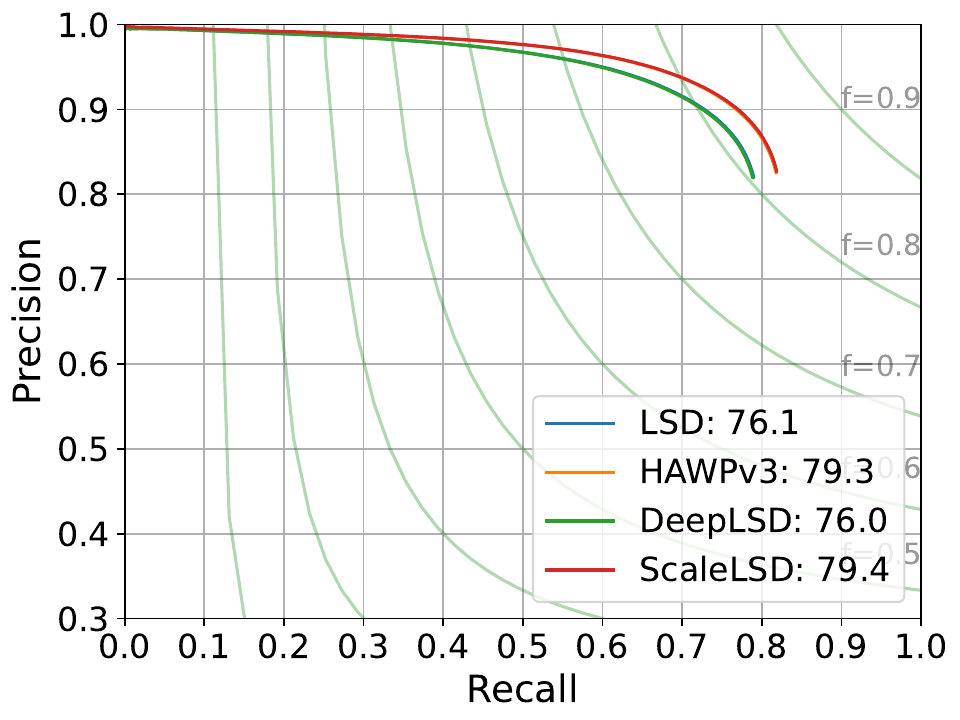}
        \caption{Points matching}
        \label{fig:pt_ap}
    \end{subfigure}
    \vspace{2mm}
    \caption{Comparison of line detectors in the performance of line matcher GlueStick \cite{gluestick} on the ETH3D dataset \cite{eth3d}. }
    \label{fig:line_match}
\end{figure}

\begin{table*}[]
    \centering
    \resizebox{\linewidth}{!}{
    \begin{tabular}{c|ccc|ccc|ccc|ccc}
    \toprule
        & \multicolumn{3}{c|}{LSD}
        & \multicolumn{3}{c|}{HAWPv3}
        & \multicolumn{3}{c|}{DeepLSD}
        & \multicolumn{3}{c}{ScaleLSD (Ours)}
        \\
        
        & ACC-L $\downarrow$ & COMP-L $\downarrow$ & \#Lines & ACC-L $\downarrow$ & COMP-L $\downarrow$ & \#Lines & ACC-L $\downarrow$ & COMP-L $\downarrow$ & \#Lines & ACC-L $\downarrow$ & COMP-L $\downarrow$ & \#Lines \\
        \cmidrule{2-4}  \cmidrule{5-7} \cmidrule{8-10} \cmidrule{11-13} 
        
scan16	&0.7043	&3.0132	&1774 &0.7898	&6.0420	&335		& 0.9242 & 2.7947 & 1957	&\textbf{0.6969}	&\textbf{2.7162}	&\textbf{2585}
\\
scan17	&0.7961	&2.3354	&2248 &0.8804	&5.8212	&388		&0.9441	&\textbf{2.2353}&2131&\textbf{0.6993}	&2.6267	&\textbf{2867}
\\
scan18	&0.8337	&2.2196	&1995 &0.8253	&7.0154	&287		& 0.9638 & \textbf{2.1534} & 1894	&\textbf{0.7357}	&2.3008	&\textbf{2563}
\\
scan19	&0.7392	&3.2416	&1424 &0.7110	&7.9461	&160	&0.9614& 3.1612& 1322 & \textbf{0.6282} & \textbf{2.4352} & \textbf{2278}
\\
scan21	&0.7890 &2.1758	&2251 &0.8884	&5.9821	&319		&0.9142 & \textbf{2.0961}& 2257	& \textbf{0.7079} & 2.4786 & \textbf{2757}
\\
scan22	&0.7808	&2.3884	&1863&0.7353	&6.8567	&281		&0.9351 & \textbf{2.2431}& 1948& \textbf{0.6593} & 2.2951 & \textbf{2442}
\\
scan24	&1.2924	&4.0612	&1213&\textbf{0.7397}	&7.7986	&246		&1.9878& \textbf{3.1395}& \textbf{1711}	&0.8366 & 4.0756 & 1624
\\\midrule
  Avg. 	&0.8479&	2.7765 &	1824&	0.7957&6.7803&	288	& 1.0901&	\textbf{2.5462}&	1888&	\textbf{0.7091}&	2.7040&	\textbf{2445}
  \\
  \bottomrule
    \end{tabular}
    }
    \caption{Quantitative results of 3D line reconstruction on the DTU~\cite{dtu} dataset for different line segment detectors.}
    \vspace{-6pt}
    \label{tab:3d_line_rec}
\end{table*}

\begin{figure*}
    \centering
    \includegraphics[width=.95\linewidth]{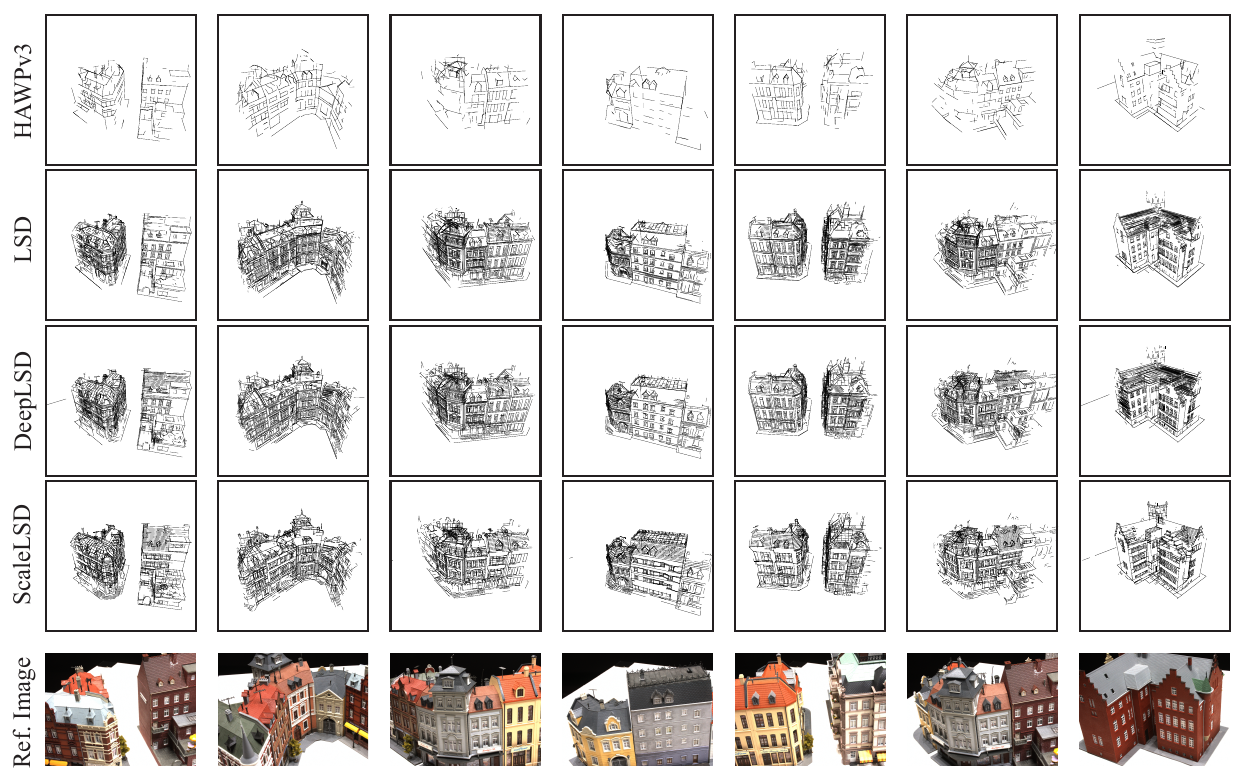}
    \caption{The qualitative comparison of 3D line mapping (by using LiMAP~\cite{limap}) with different line segment detectors on the building scenes of the DTU dataset~\cite{dtu}. The video results can be found at our project page.
    }
    \vspace{-6pt}
    \label{fig:appx-dtu}
\end{figure*}

\subsection{Line Matching Evaluation}
Good line segment detectors are always expected for two-view line segment matching. In this experiment, we feed the detection results into the state-of-the-art line matcher, GlueStick~\cite{gluestick} to yield matches from two-view images. The comparisons are made on the ETH3D dataset~\cite{eth3d} and we use the precision, recall and F1-score for the resulted point and line matches as the metrics.

\vspace{-15pt}
\paragraph{Protocol and Results.} 
Because our method focus on the detection, we build the matcher by using SuperPoint~\cite{SuperPoint} as the feature descriptor of junction (or endpoint of line segments) for different detectors. We show the matching precision-recall curves of lines (~\cref{fig:line_ap}) and points (~\cref{fig:pt_ap}, consists of keypoints and junctions/endpoints) in ~\cref{fig:line_match}, and we also attach the average precision (AP) value after each line detector tag of these two sub-figures. It is obviously that our method has significantly exceeded gradient-based methods LSD and DeepLSD and outperforms HAWPv3 greatly in lines matching. For the visualization of line segment matching on challenging cases, please refer to our supplementary material.

\subsection{3D Line Reconstruction}\label{sec:3dline}
Based on the aforementioned experimental results on detection repeatability, vanishing points estimation and line matching, we move forward to multi-view 3D line reconstruction to evaluate the performance of our \method. 

\vspace{-10pt}
\paragraph{Protocol and Metrics.} The line mapping framework~\cite{limap} is used in our experiments, which follows a pipeline that sequentially (1) detect line segments and estimate vanishing points from images, (2) match line segments and build line tracks and VP tracks as well, (3) triangulate the line tracks into 3D space using the given camera parameters. 
7 scenes of building from the DTU~\cite{dtu} dataset are used for evaluation, in which we compute the ACC and COMP errors between the predicted line segments and the GT point clouds provided by the dataset for each scene. Four detectors, LSD~\cite{TP-LSD}, HAWPv3~\cite{HAWPv3}, DeepLSD~\cite{DeepLSD} and our method are evaluated. 
We additionally report the number of reconstructed 3D line segments as reference for comparison.
The detailed evaluation protocol is deferred to supplementary material.

\vspace{-10pt}
\paragraph{Results.} \cref{tab:3d_line_rec} reports the quantitative evaluation results on the DTU~\cite{dtu} dataset for LiMAP with different detectors. Compared with other detectors, our method obtains the best ACC scores while keeping reasonable completeness for the 3D line reconstruction. \cref{fig:appx-dtu} visualizes the reconstruction results. 

\begin{table}[!h]
    \centering
    \resizebox{\linewidth}{!}{
    \begin{tabular}{ccccccccc}
    \toprule
    LSD- & Homo. & \multicolumn{2}{c}{Struct} & \multicolumn{2}{c}{Orth} & $\#$~Lines & Avg & Mem \\
    Rectifier & Adapt. & Rep5~$\uparrow$ & Loc5~$\downarrow$ & Rep5~$\uparrow$ & Loc5~$\downarrow$ & /~Image & Time[s] & [MB] \\
    \midrule
         & & 0.397 & 2.521 & 0.562 & 1.688 & \textbf{80} & \textbf{0.381} & \textbf{6574}  \\
         & \checkmark & \underline{0.447} & 2.452 & 0.574 & 1.617 & 28 & 3.653 & 45378  \\ 
        \checkmark& & 0.445 & \textbf{2.377} & \textbf{0.630} & \underline{1.602} & \underline{78} & \underline{0.607} & \underline{6588}  \\
         \checkmark & \checkmark & \textbf{0.473} & \underline{2.444} & \underline{0.621} & \textbf{1.592} & 32 & 5.811 & 45492 
    \\\bottomrule
    \end{tabular}
    }
    \caption{The ablation study of using our LSD-Rectifier and classical Homographic Adaptation for pseudo label generation on the Wireframe dataset. We report the metrics and line numbers to compare the effectiveness and report the average time and space overhead for one batch to compare the efficiency.}
    \label{tab:abl_labels}
\end{table}
\vspace{-8pt}

\begin{table}[!h]
    \centering
    \resizebox{\linewidth}{!}{
    \begin{tabular}{cccccccc}
    \toprule
     \multicolumn{2}{c}{Verification} & \multicolumn{2}{c}{Backbone} & \multicolumn{2}{c}{Struct} & \multicolumn{2}{c}{Orth} \\
     LOI-based & HAT-induced & Hourglass & DPT & Rep-5 (S) $\uparrow$ & Loc-5 (S) $\downarrow$ & Rep-5 (O) $\uparrow$ & Loc-5 (O) $\downarrow$ \\
    \midrule
    \checkmark & & \checkmark & & 0.356 & 2.912 & 0.629	& 1.890 \\
    \checkmark & & & \checkmark & 0.418 & 2.763	& \textbf{0.643} & 1.856 \\
    & \checkmark & \checkmark & & 0.263&	2.752	&0.584	&1.852 \\
    & \checkmark & & \checkmark & \textbf{0.445} & \textbf{2.377} & 0.630 & \textbf{1.602} \\
    \bottomrule
    \end{tabular}
    }
    \caption{The ablation study of using different verifications and backbones for the synthetic bootstrapping stage on the Wireframe dataset.}
    \label{tab:abl_bp}
\end{table}
\vspace{-8pt}

\begin{table}[!h]
    \centering
    \resizebox{\linewidth}{!}{
    \begin{tabular}{r|cccccccccc|c}
    \toprule
     & \multicolumn{10}{c|}{Homo. Adapt.} & LSD- \\ \cline{2-11}
     \textit{iter\_num} & 1 & 2 & 3 & 4 & 5 & 6 & 7 & 8 & 9 & 10 & Rectifier \\
    \midrule
    $\#$~Lines/Image & 35 & 27 & 39 & 33 & 30 & 28 & 33 & 30 & 29 & 28 & 78 \\
    Avg Time[s] & 0.702 & 0.944 & 1.371 & 1.562 & 1.973 & 2.388 & 2.693 & 2.945 & 3.293 & 3.653 & 0.607 \\
    Mem[MB] & 10588 & 14400 & 18214 & 22042 & 25856 & 29956 & 33812 & 37666 & 41522 & 45378 & 6588 \\
    \bottomrule
    \end{tabular}
    }
    \caption{The ablation study of using different number of iteration for Homographic Adaptation for pseudo label generation on the Wireframe dataset.}
    \label{tab:abl_homo}
\end{table}

\subsection{Ablation Study}
We verify our main designs for line segment detection from two aspects, including the verification of line proposals (~\cref{subsec:proposal-verification}) and the generation of pseudo labels (~\cref{subsec:Pseudo-Label-Generation}).

\vspace{-10pt}
\paragraph{Line Proposals Verification}
We compare our proposed HAT-induced Proposal Verification with the classical LOI-based verification scheme by testing the learned synthetic models on a hybrid dataset, containing 2,000 images randomly sampled from the Wireframe dataset, the SA1B dataset and the HPatches dataset. 
We further make the discussion about the impact of CNN-based and transformer-based backbones for the detection performance of these two line proposals verifications.
As shown in ~\cref{tab:abl_bp}, compared to HAWPv3~\cite{HAWPv3} which uses the LOI-based verification scheme, our \method achieves better results in all metrics, demonstrating the effectiveness of the proposed HAT-induced Proposal Verification. 
Additionally, LOI-based verification shows negligible scalability of its architecture as the scale-up DPT makes limited improvement of performance relative to the small Hourglass. In contrast, our HAT-induced verification applied with DPT makes significant improvement compared with ones of Hourglass, which shows its promising scalability on applying bigger and powerful backbone for LSD.

\vspace{-15pt}
\paragraph{Pseudo Labels Generation}
As discussed in ~\cref{subsec:Pseudo-Label-Generation}, we use LSD-Rectifier for Pseudo Label Generation instead of the commonly used homographic adaptation (Homo.Adap.) scheme~\cite{SOLD2,SuperPoint}. We use the trained synthetic model to compare these two schemes by evaluating the quality of their generated pseudo labels on the Wireframe dataset~\cite{wireframe-dataset}. As shown in ~\cref{tab:abl_labels}, our LSD-Rectifier strategy achieves comparable repeatability score and localization error, while generating much more line segments than `Homo. Adap.', which is important for subsequent learning. Besides, our LSD-Rectifier is much faster than `Homo. Adap.' and is more suitable for large-scale data generation. We set the score threshold for homographic adaptation to 0.75. We also make the ablation study about the impact of iteration number to detected lines number during homographic adaptation in ~\cref{tab:abl_homo}.

\section{Conclusion}\label{sec:conclusion}
This paper addressed the problem of line segment detection in self-supervised learning. To tackle the generalization issues persisting in current approaches, typically trained on small-scale datasets of about 20k images, we developed the first model trained using 10M unlabeled data. In designing our method, we critically evaluated prevailing designs, spanning from classical LSD to the recently proposed HAT field representation, streamlining the entire learning pipeline with simple and intuitive designs. Leveraging the powerful scalability inherent in Transformers, we have successfully achieved our goal of generalizable and data-driven line segment detection. This achievement has been demonstrated through various evaluation protocols, including cross-view repeatability, vanishing point estimation, line segment matching and 3D line reconstruction, where we surpassed state-of-the-art performance benchmarks. 
We believe that our study, which focuses on the symbolic representation of boundary geometry in images, has the potential to offer a parsimonious representation of visual data using a small number of primitives.

\paragraph{Limitations}
While our method addresses the generalization problem in learning-based line segment detection by utilizing a significantly larger scale of unlabeled data (10M images) compared to prior approaches, we did not fully explore its scalability potential with even larger datasets. Consequently, there remains a risk of under-detecting line segments in testing images. Additionally, while the powerful generalization ability of our method could characterize curves in polylines, our method does not explicitly take curve structures into the modeling process.

\section*{Acknowledgement}
This work was supported by Ant Group Research Intern Program.

{
\small
\bibliographystyle{ieeenat_fullname}
\bibliography{ref.bib}
}

\onecolumn
\appendix
\section*{Appendix}

\begin{figure}[htbp]
    \centering
    \begin{subfigure}[b]{0.3\linewidth}
        \centering
        \includegraphics[width=\linewidth]{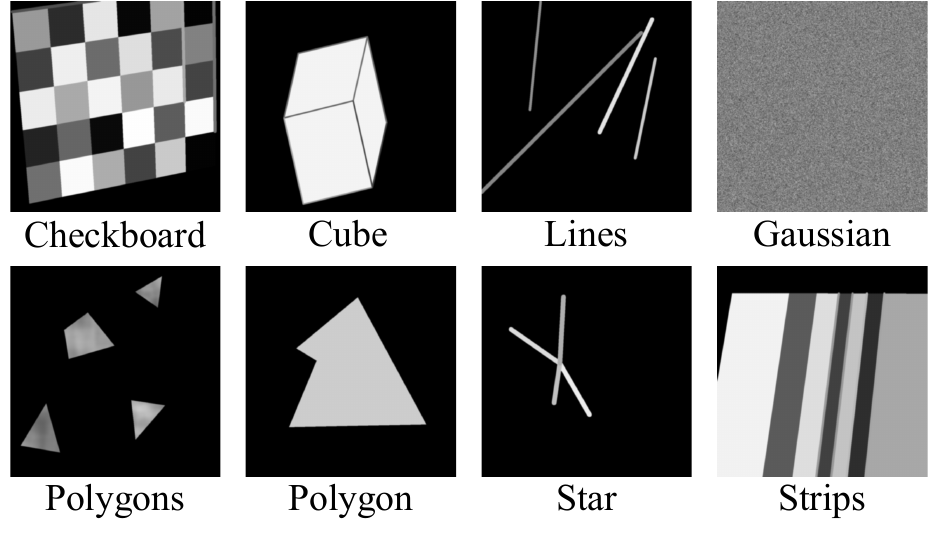}
        \caption{Synthetic data examples}
        \label{fig:data_syn}
    \end{subfigure}
    \hfill
    \begin{subfigure}[b]{0.65\linewidth}
        \centering
        \includegraphics[width=\linewidth]{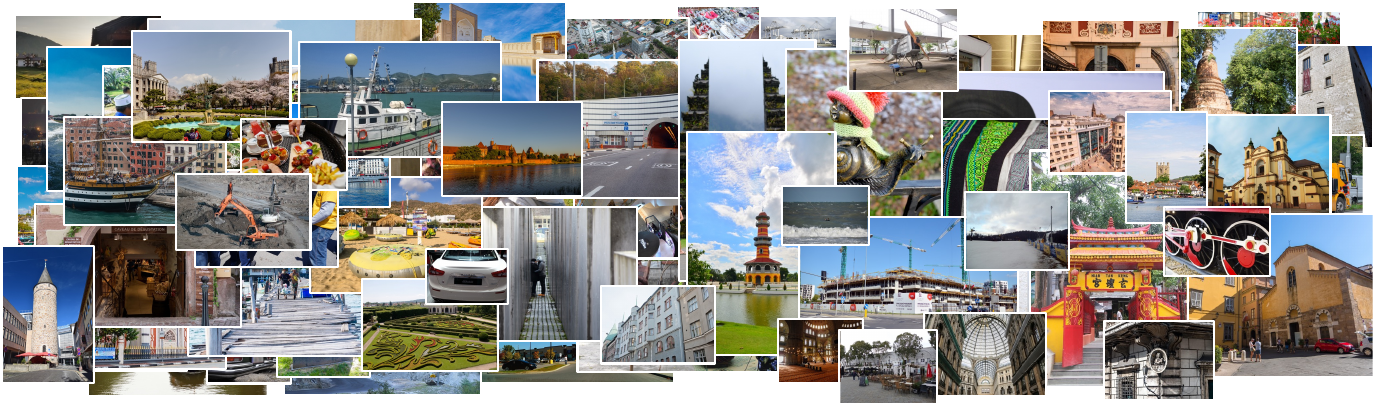}
        \caption{Real data examples}
        \label{fig:data_real}
    \end{subfigure}
    \caption{Some training examples on the generated synthetic dataset and the real SA1B~\cite{SAM} dataset.}
    \label{fig:data-sa1b}
\end{figure}

\section{Additional Implementation Details}
\paragraph{Training Data and Pipelines.}
Our training pipelines are similar with previous studies~\cite{SuperPoint,SOLD2,HAWPv3}. In the  bootstrapping stage learns the concept of line segments from the synthetic images using 8 simple primitives as shown in \cref{fig:data_syn}. With the bootstrapping model, we move forward to the small-scale Wireframe~\cite{wireframe-dataset} dataset to learn the line segments in real-world images, and use this model to achieve the largest-scale training of LSD on the SA-1B~\cite{SAM} dataset, which contains 10 million real-world image samples as shown in \cref{fig:data_real}.

\paragraph{Network Architecture.}
Our network architecture is simple, follows the best practices of vision transformers for dense predictions~\cite{DPT}. In detail, given a batch $B$ of RGB images with shape $512\times 512$, a ViT-Base model is used to extract $1024$ tokens for dense prediction of HAT fields and junction heatmaps. DPT head is applied to first transform the $1024$ tokens into high-resolution feature maps with the shape of $[B\times N\times 256 \times 256]$, and then predict the HAT fields and junction heatmaps using $1\times 1$ convolution layers.
In our model, there are no neural modules for the verification of line segments, which has greatly simplified the training and inference pipeline compared to HAWPv3~\cite{HAWPv3}.

\paragraph{Loss Functions.}
We use the $\mathcal{L}_1$ loss function for the regression of the distance field $\mathcal{A}_d$, the angle field $\mathcal{A}_a$ and the residual distance $\mathcal{A}_{\Delta d}$, denoted by $(\mathcal{L}_d, \mathcal{L}_a, \mathcal{L}_{\Delta d})$. The loss is computed across the foreground points only based on the mask of foreground pixels.
We use binary cross-entropy loss ${\rm BCE}(\cdot,\cdot)$ for the regression of the endpoints and use loss $\mathcal{L}_1$ for the regression of the offset field, record as $(\mathcal{L}_j, \mathcal{L}_o)$.
We set the weights of each loss to $(\lambda_d, \lambda_a, \lambda_{\Delta d}, \lambda_j, \lambda_o) = (1.0, 1.0, 1.0, 8.0, 0.25)$, and the total loss of our model is
\begin{equation}
    \mathcal{L} = \underbrace{\lambda_d \mathcal{L}_d + \lambda_a \mathcal{L}_a + \lambda_{\Delta d} \mathcal{L}_{\Delta d}}_{\text{HAT Field Learning}} +\overbrace{\lambda_j \mathcal{L}_j  + \lambda_o \mathcal{L}_o}^{\text{Junction Learning}}.
\end{equation}
The setting of $\lambda_j$ and $\lambda_o$ follows HAWPv3~\cite{HAWPv3} to balance the significant magnitude difference of these two loss terms.

\paragraph{Inference.}
Our \method takes any RGB/grayscale image as input, predicts the HAT fields and junction heatmaps using a neural network, and decodes the hat fields and junction heatmaps into sparse line segments. In the decoding stage, the junction heatmaps are first processed by a max-pooling layer with a window size of $3$ to suppress the non-maximal predictions, then we extract the top-$K$ pixels as the coarse junction predictions. When the junction score (\ie, heatmap value) of any pixel is less than $\tau_j\in(0,1)$, it is discarded. The junction score threshold $\tau_j$ is set to 0.008 for training and pseudo-label generation and is set to 0.1 for inference and evaluation. For the finally-kept coarse junctions, we apply the learned short-range offset to obtain final junctions with sub-pixel localization accuracy.
With the extracted junction, we decode the line segments by matching them to the line segment fields (computed by the HAT fields) according to Eq.~(2) of our main paper. The distance threshold $\tau_{\rm dist}$ is set to 10 pixels, rejecting low-quality predictions in the HAT fields from the final predictions. By matching the junction to lines, the line segments whose support pixels are larger than $\tau_l$ are kept as the final predictions. Here, we set $\tau_l$ to 10 for training and pseudo-label generation and is set to 5 for inference and evaluation.

\paragraph{Details on 3D Line Reconstruction.}
In 3D line reconstruction, we found the threshold of top-$K$ should be increased to $2048$ because the buildings usually have more structural information. 
For the evaluation, we follow the protocol provided by DTU dataset~\cite{dtu} to compute the Chamfer distance between the predicted line segments (sampled in 128 points per line) and the groundtruth surface model. The accuracy (ACC-L) and the completeness (COMP-L) are computed to measure the reconstruction quality. We also add the number of 3D line segments as a reference. 
We reconstruct the 3D lines using LiMAP~\cite{limap} by switching the line segment detectors. The line matching module is their built-in GlueStick~\cite{gluestick} implementation for all detectors.

\section{Visualization of VP Estimation}

\begin{figure*}[!h]
    \centering
        \includegraphics[width=.9\linewidth]{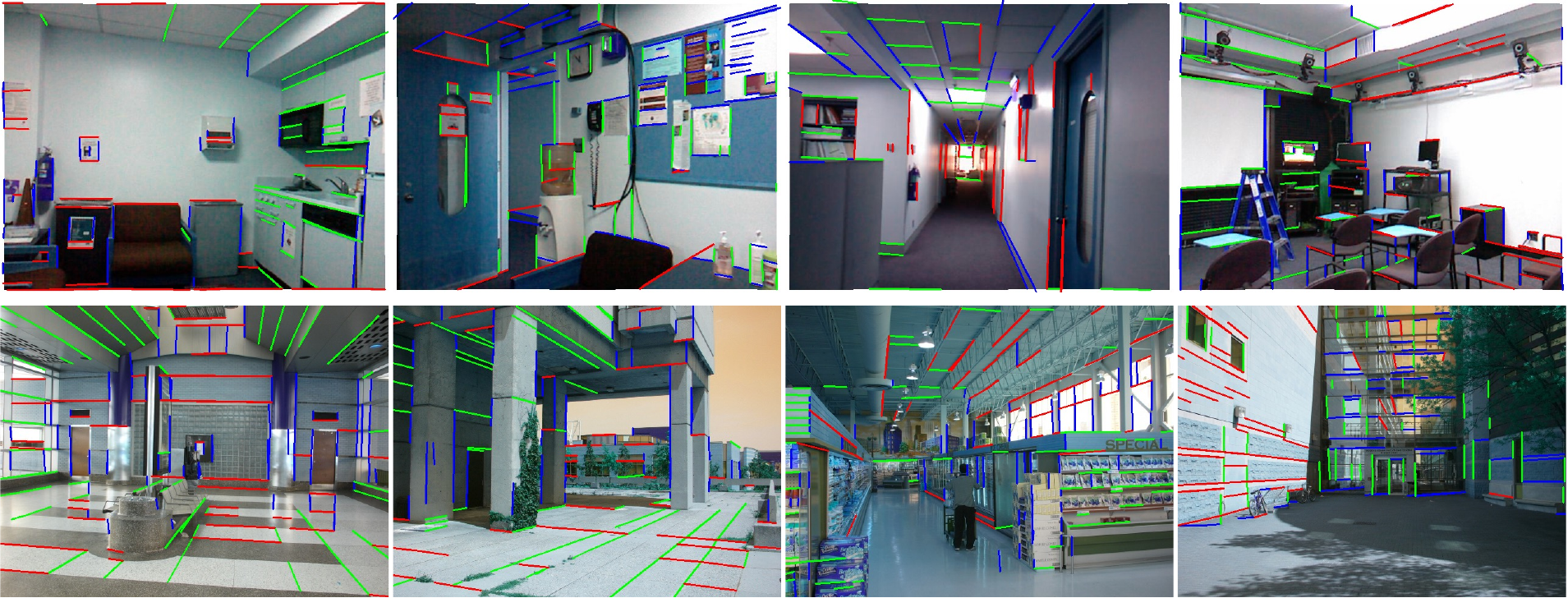}
    \caption{The illustration of vanishing points estimation. Lines belong to the same one vanishing point are labeled with the same color. Top row shows the results of the Manhattan scenes in the YUD+~\cite{YUD+} dataset and bottom row shows the results of non-Manhattan scenes in the NYU-VP~\cite{NYU-VP} dataset.}
    \label{fig:vps}
\end{figure*}

We visualize results of vanishing points estimation by drawing the parallel line segments associated with different vanishing points in different colors. \cref{fig:vps} shows that, our method could robustly estimate vanishing points in both Manhattan and Non-Manhattan scenes. To better show the results, the line segments that are not associated with any vanishing points are hidden to display.

\section{Visualization of Line Matching}
\begin{figure*}[!h]
    \centering
        \includegraphics[width=.9\linewidth]{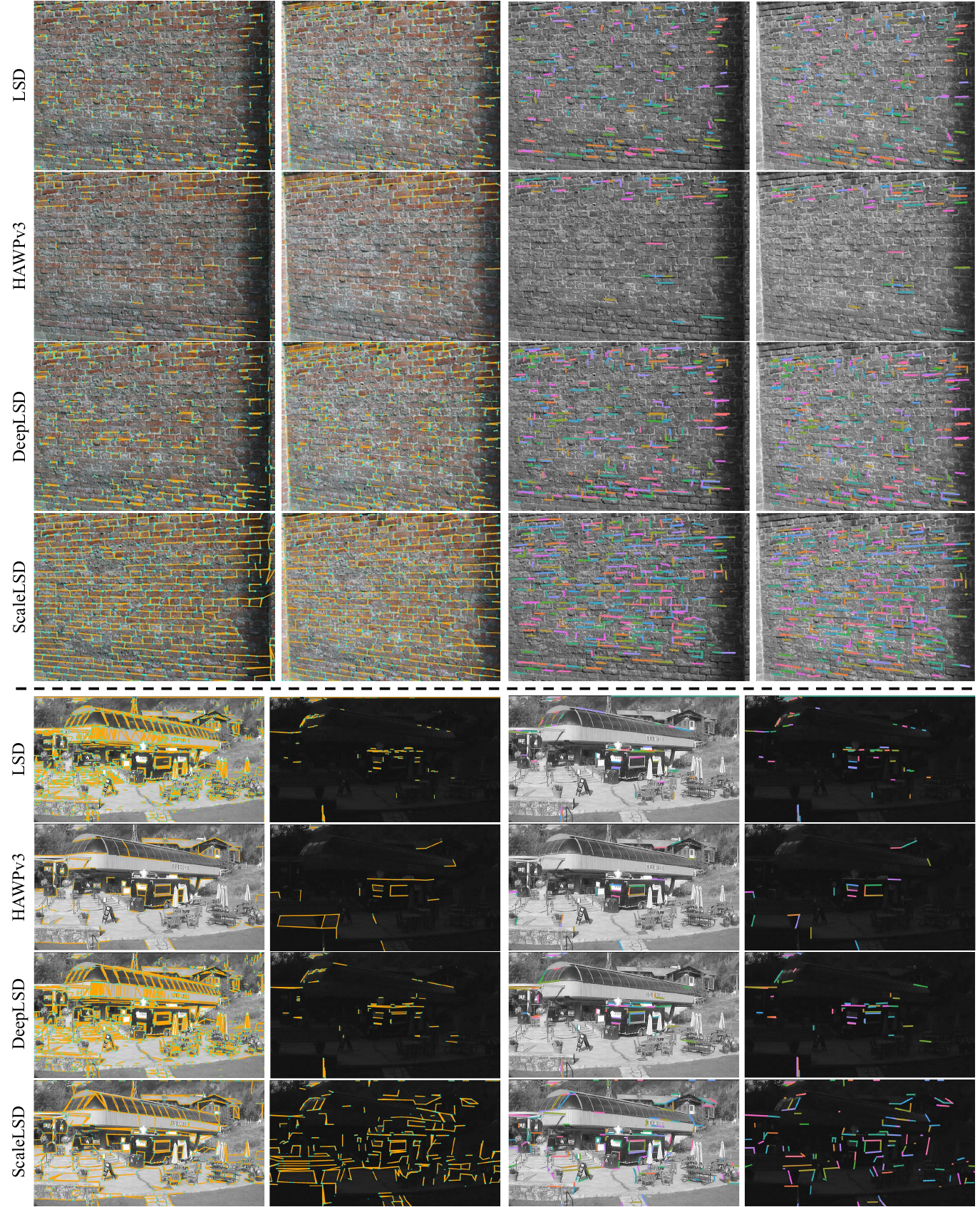}
    \caption{Challenging examples of line segment matching. For each case, from left to right, we first show the detection results for the two-view input images, and then show the matched line segments.
    Top: we show the challenge pair of images that have similar structure and texture as well as change of viewpoint. Bottom: we show the challenge pair of images that have significant illumination changes. Lines with the same color in the last two images means the matched pairs.}
    \label{fig:lm_vis}
\end{figure*}

Line segments matching is a challenge task due to common situations of changes of view and illumination, occlusions, background changes, repeatable structures, and textures. 
Two typical challenging cases for repeatable structures and intensive illumination changes between the input image pairs are shown in \cref{fig:lm_vis}. As shown, because our \method significantly improves detection performance in terms of detection completeness, the applied line segment matcher (\ie, GlueStick~\cite{gluestick}) could leverage the global information conveyed in the structural line segment representation for better matching.

\begin{table*}[!h]
    \centering
    \resizebox{\linewidth}{!}{
    \begin{tabular}{r|ccccc|ccccc}
    \toprule
\multirow{2}{*}{Method} & \multicolumn{5}{c|}{Wireframe} & \multicolumn{5}{c}{SA1B-1000} \\
& Rep-5 (S) $\uparrow$ & Loc-5 (S) $\downarrow$ &	Rep-5 (O) $\uparrow$ & Loc-5 (O) $\downarrow$&	\#Lines/Image&	Rep-5 (S) $\uparrow$ & Loc-5 (S) $\downarrow$ &	Rep-5 (O) $\uparrow$ &	Loc-5 (O) $\downarrow$ &		\#Lines/Image\\
\midrule
LSD~\cite{VonGioi2010} & 0.383 & 2.198 & 0.719 & 1.028 & \underline{441} & 0.432 & 2.179 & 0.665 & 1.153 & \textbf{614} \\
SOLD$^2$~\cite{SOLD2} & 0.566 & 2.039 & 0.805 & 1.135 & 116 & 0.480 & 2.226 & 0.688 & 0.954 & 97 \\
HAWPv3~\cite{HAWPv3} & \textbf{0.751} & \underline{1.487} & \textbf{0.874} & \underline{0.841} & 145 & 0.519 & \underline{1.680} & 0.664 & \textbf{0.905} & 125 \\
DeepLSD~\cite{DeepLSD} & 0.512 & 2.236 & 0.707 & 1.085 & 210 & 0.396 & 2.400 & 0.601 & 1.265 & 181 \\
ScaleLSD@Wireframe(Ours) & 0.723 & 1.694 & \underline{0.822} & 0.897 & 413 & \underline{0.555} & 1.856 & \underline{0.692} & 0.955 & 419 \\
ScaleLSD@SA1B(Ours) & \underline{0.725} & \textbf{1.466} & 0.820 & \textbf{0.837} & \textbf{764} & \textbf{0.634} & \textbf{1.535} & \textbf{0.728} & \underline{0.911} & \underline{580} 
\\\bottomrule
    \end{tabular}
    }
    \caption{ Evaluation of repeatability scores and localization errors on in-domain datasets. The image resolution are fixed to 512$\times$512 in evaluation. Numbers with \textbf{bold-font} and \underline{underline} indicate the best and the second best performance on specific metrics.}
    \label{tab:ind}
\end{table*}

\begin{figure*}
    \centering
    \begin{subfigure}{\linewidth}
        \includegraphics[width=\linewidth]{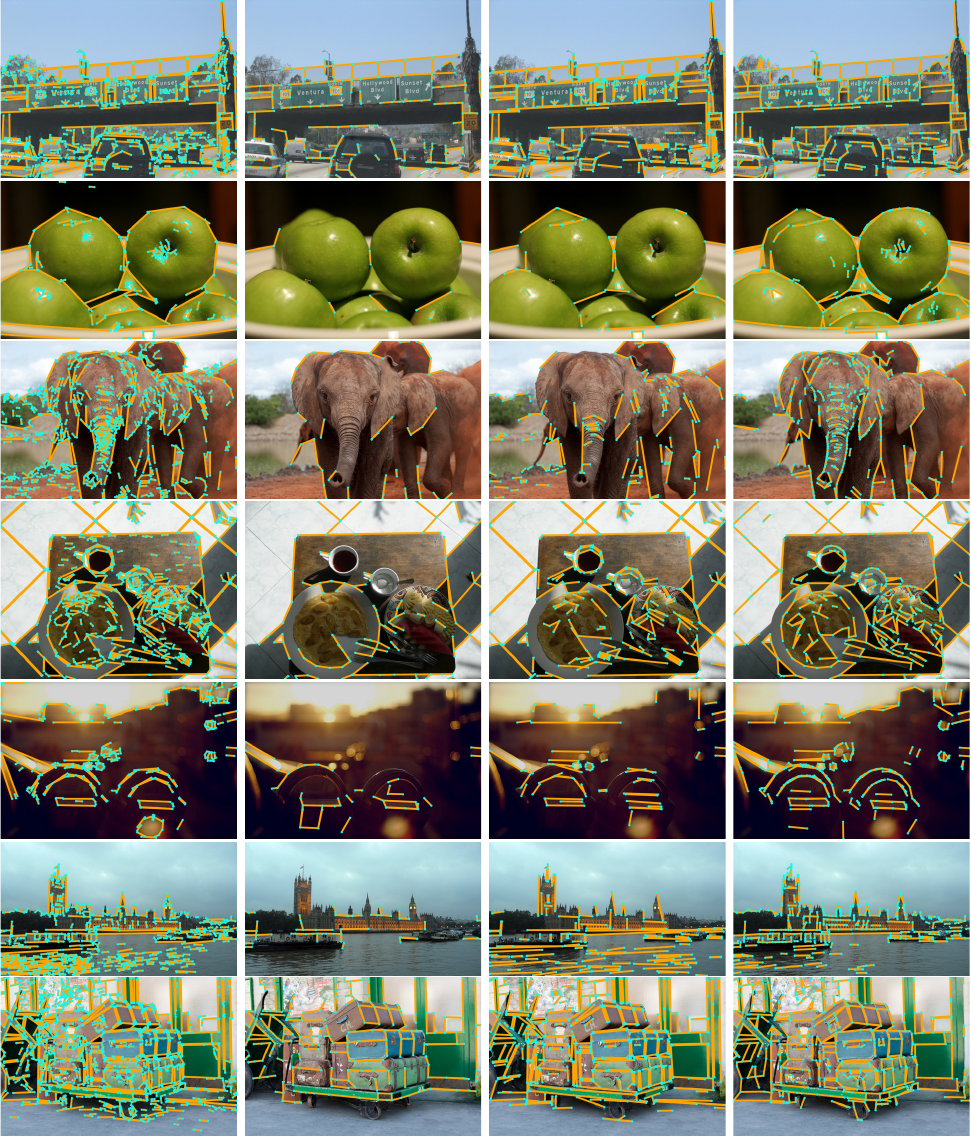}
    \end{subfigure} \hfill
    \begin{minipage}{0.24\textwidth}
        \centering LSD~\cite{LSD}
    \end{minipage}
    \begin{minipage}{0.24\textwidth}
        \centering HAWPv3~\cite{HAWPv3}
    \end{minipage}
    \begin{minipage}{0.24\textwidth}
        \centering LSD~\cite{DeepLSD}
    \end{minipage}
    \begin{minipage}{0.24\textwidth}
        \centering ScaleLSD
    \end{minipage}
    \caption{Qualitative results of line segments detection.}
    \label{fig:vis1}
\end{figure*}
\begin{figure*}
    \centering
    \begin{subfigure}{.95\linewidth}
        \includegraphics[width=\linewidth]{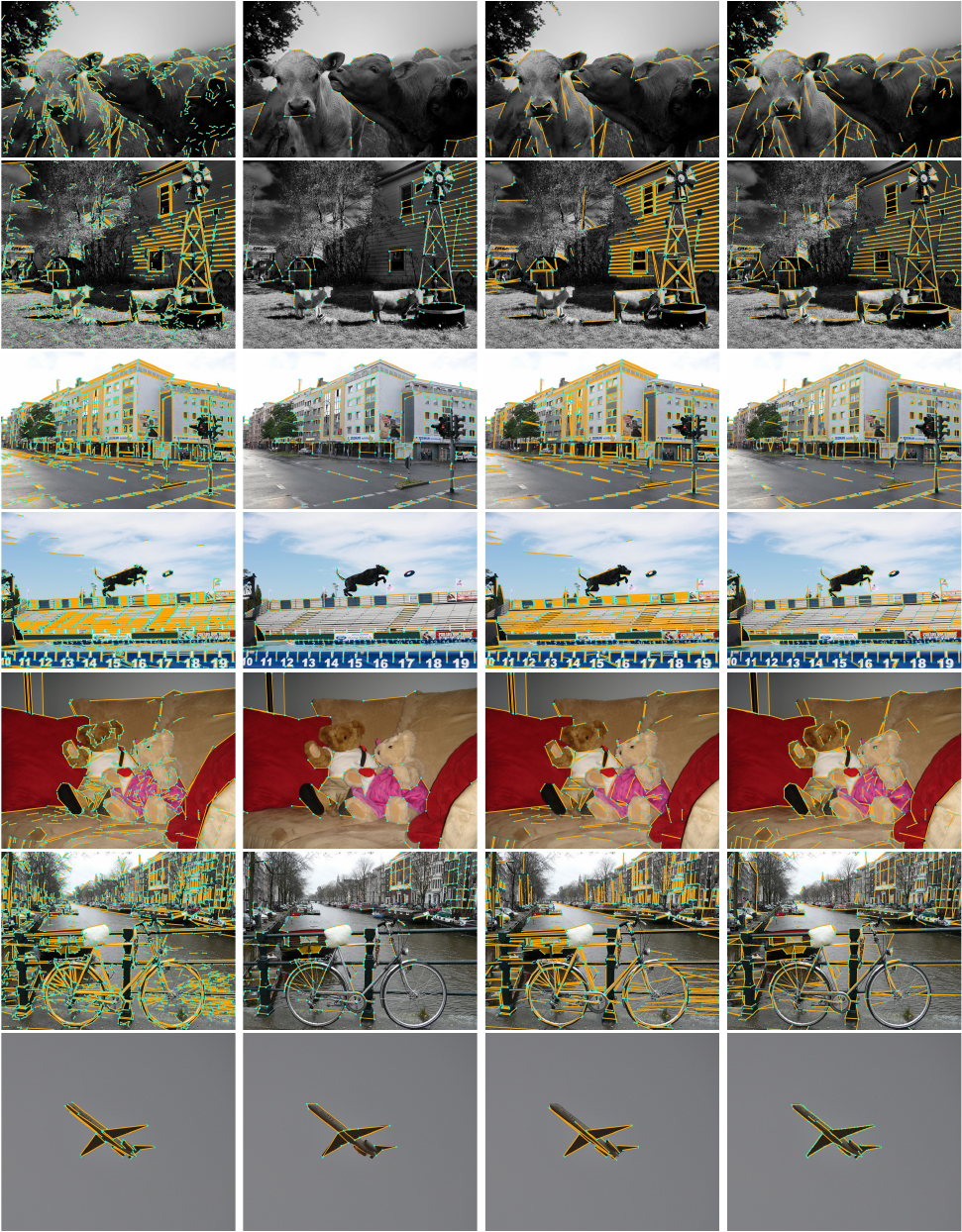}
    \end{subfigure} \hfill
    \begin{minipage}{0.24\textwidth}
        \centering LSD~\cite{LSD}
    \end{minipage}
    \begin{minipage}{0.24\textwidth}
        \centering HAWPv3~\cite{HAWPv3}
    \end{minipage}
    \begin{minipage}{0.24\textwidth}
        \centering DeepLSD~\cite{DeepLSD}
    \end{minipage}
    \begin{minipage}{0.24\textwidth}
        \centering ScaleLSD
    \end{minipage}
    \caption{Qualitative results of line segments detection.}
    \label{fig:vis2}
\end{figure*}
\begin{figure*}
    \centering
    \begin{subfigure}{\linewidth}
        \includegraphics[width=\linewidth]{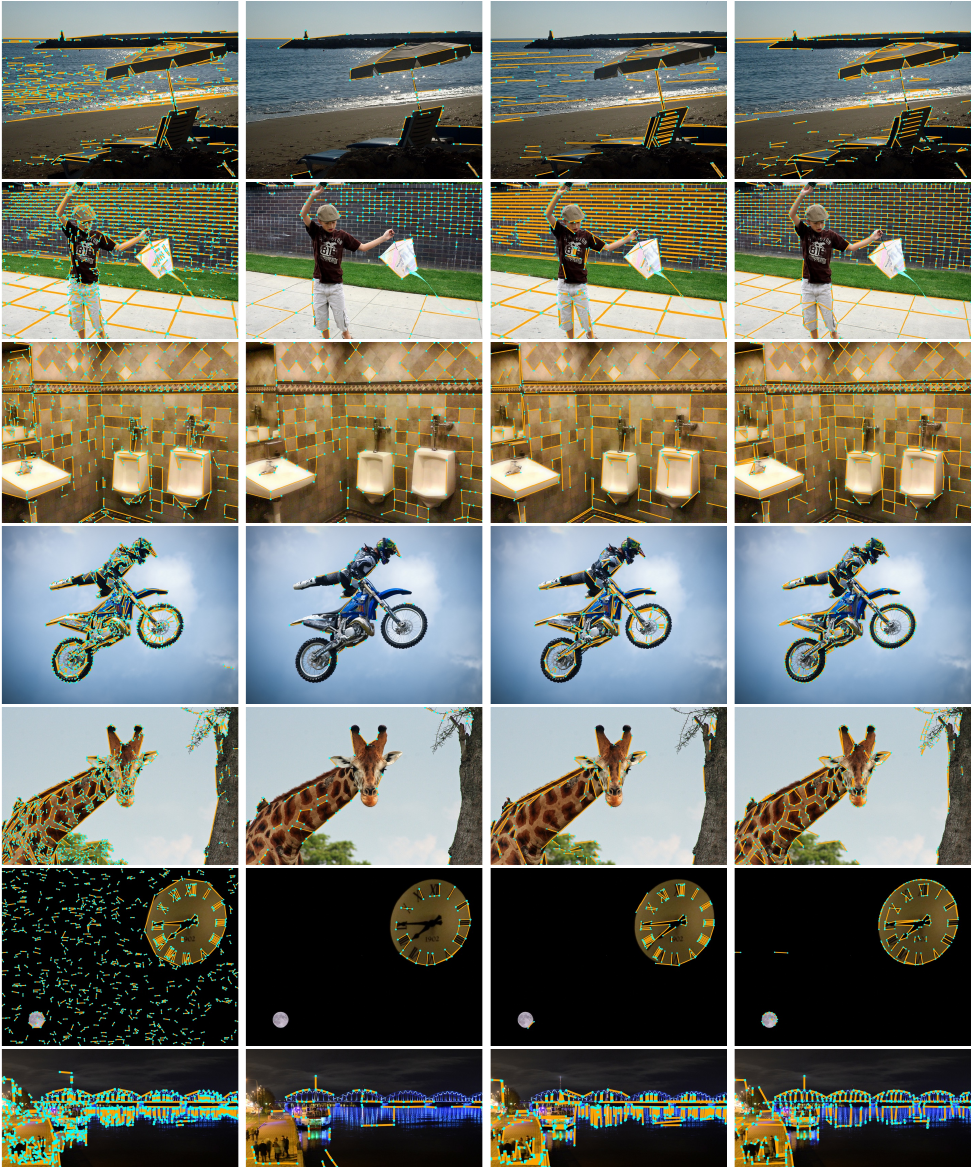}
    \end{subfigure} \hfill
    \begin{minipage}{0.24\textwidth}
        \centering LSD~\cite{LSD}
    \end{minipage}
    \begin{minipage}{0.24\textwidth}
        \centering HAWPv3~\cite{HAWPv3}
    \end{minipage}
    \begin{minipage}{0.24\textwidth}
        \centering DeepLSD~\cite{DeepLSD}
    \end{minipage}
    \begin{minipage}{0.24\textwidth}
        \centering ScaleLSD (Ours)
    \end{minipage}
    \caption{Qualitative results of line segments detection.}
    \label{fig:vis3}
\end{figure*}
\begin{figure*}
    \centering
    \begin{subfigure}{\linewidth}
        \includegraphics[width=\linewidth]{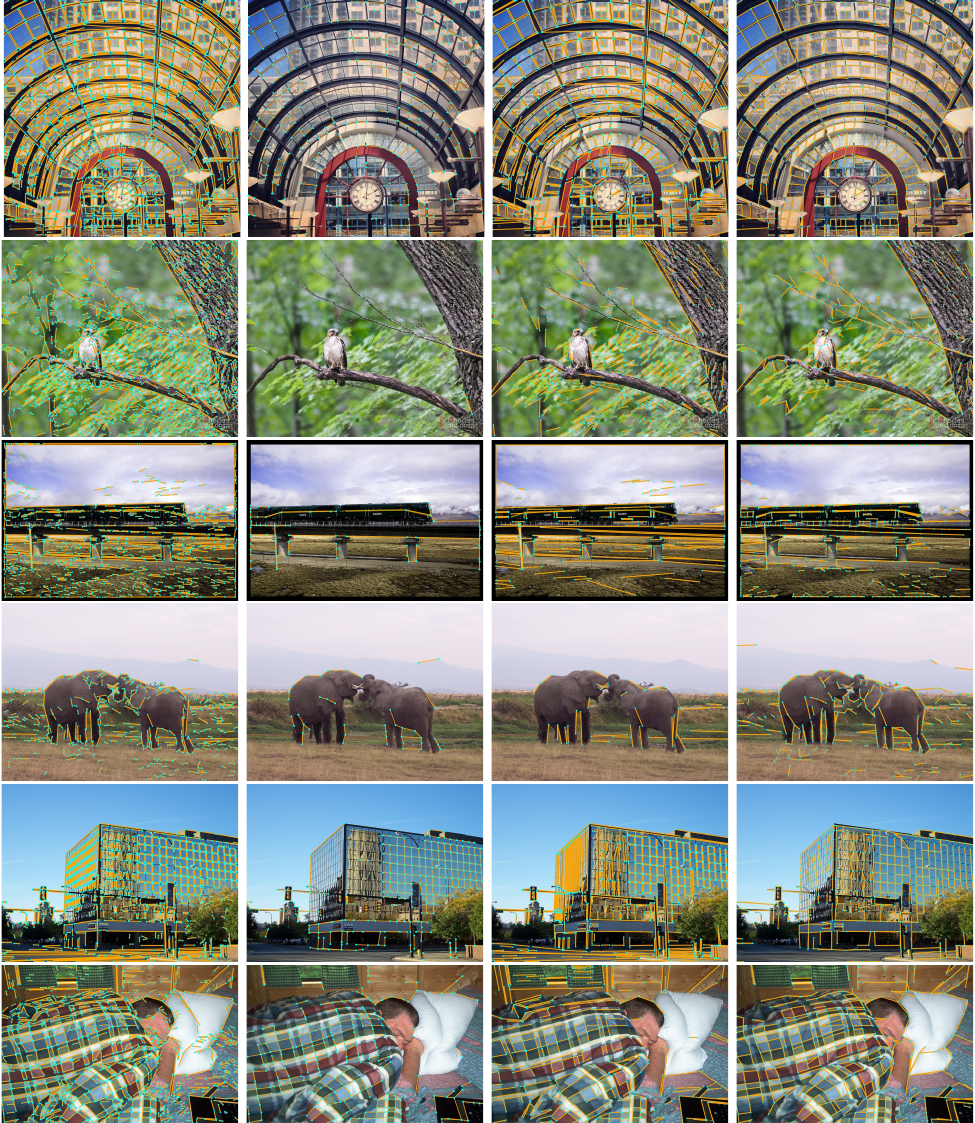}
    \end{subfigure} \hfill
    \begin{minipage}{0.24\textwidth}
        \centering LSD~\cite{LSD}
    \end{minipage}
    \begin{minipage}{0.24\textwidth}
        \centering HAWPv3~\cite{HAWPv3}
    \end{minipage}
    \begin{minipage}{0.24\textwidth}
        \centering DeepLSD~\cite{DeepLSD}
    \end{minipage}
    \begin{minipage}{0.24\textwidth}
        \centering ScaleLSD (Ours)
    \end{minipage}
    \caption{Qualitative results of line segments detection.}
    \label{fig:vis4}
\end{figure*}

\end{document}